\documentclass[letterpaper, 10 pt, conference]{ieeeconf}

\IEEEoverridecommandlockouts                              

\usepackage{dblfloatfix} 
\usepackage{balance}
\usepackage{graphics}
\usepackage{cite}
\usepackage{amsmath,amssymb,amsfonts}
\usepackage{algorithmic}
\usepackage{textcomp}
\usepackage{courier}
\usepackage{float} 
\usepackage{threeparttable}
\usepackage{array}
\usepackage{multirow}
\usepackage{longtable}
\usepackage{rotating}
\usepackage{academicons}
\usepackage{setspace}
\usepackage{makecell}
\IEEEpeerreviewmaketitle
\usepackage{lipsum}
\begin{document}
\title{\LARGE \bf Design of a Flying Humanoid Robot Based on Thrust Vector Control}
\author{Yuhang Li$^{\dag}$, Yuhao Zhou$^{\dag}$, Junbin Huang, Zijun Wang, Shunjie Zhu, Kairong Wu, Li Zheng, Jiajin Luo, \\Rui Cao, Yun Zhang,~\IEEEmembership{Member,~IEEE}, and Zhifeng Huang$^{*}$,~\IEEEmembership{Member,~IEEE}
\thanks{$^{\dag}$
    These authors contribute equally to this work.}
    \thanks{$^{*}$
    Corresponding author: Zhifeng Huang}
\thanks{This work was supported by the National Natural Science Foundation of China (NSFC) under Grant No. 51605098, and in part by the Natural Science Foundation of Guangdong Province, China under Grant 2021A1515011829.}
\thanks{The authors are with the School of Automation, Guangdong
University of Technology, Guangzhou 510006, P. R. China. {\tt\small{zhifeng@gdut.edu.cn}}}
}
\maketitle
\thispagestyle{empty}
\pagestyle{empty}

\begin{abstract}
Achieving short-distance flight helps improve the efficiency of humanoid robots moving in complex environments (e.g., crossing large obstacles or reaching high places) for rapid emergency missions. This study proposes a design of a flying humanoid robot named Jet-HR2 (Fig. \ref{fig:Fig1}). The robot has 10 joints driven by brushless motors and harmonic drives for locomotion. To overcome the challenge of the stable-attitude takeoff in small thrust-to-weight conditions, the robot was designed based on the concept of thrust vectoring. The propulsion system consists of four ducted fans, that is, two fixed on the waist of the robot and the other two mounted on the feet for thrust vector control. The thrust vector is controlled by adjusting the attitude of the foot during the flight. A simplified model and control strategies are proposed to solve the problem of attitude instability caused by mass errors and joint position errors during takeoff. The experimental results showed that the robot’s spin and dive behaviors during takeoff were effectively suppressed by controlling the thrust vector of the ducted fan on the foot. The robot successfully achieved takeoff at a thrust-to-weight ratio of 1.17 (17 kg / 20 kg) and maintained a stable attitude, reaching a takeoff height of over 1000 mm.
\end{abstract}
\section{Introduction}
Recently, various disaster-response humanoid robots have been invented with unique control theories and other mechanisms to overcome uneven terrain. Traditionally, humanoid robots have overcome these obstacles by stepping \cite{kaneko2015humanoid, jung2018development} and climbing \cite{yoshiike2017development, yoshiike2019experimental}, yet these strategies lack efficiency, especially for dangerous environments like insurmountable obstacles and geological faults. For urgent tasks in complex real scenarios, humanoid robots are expected to have dynamic aerial skills, such as high or long jumps, short-distance flights, and hovering that exceed the body length several times.
\begin{figure}[t]
\vspace{0.2cm}
\centerline{\includegraphics[width=0.326\textwidth]{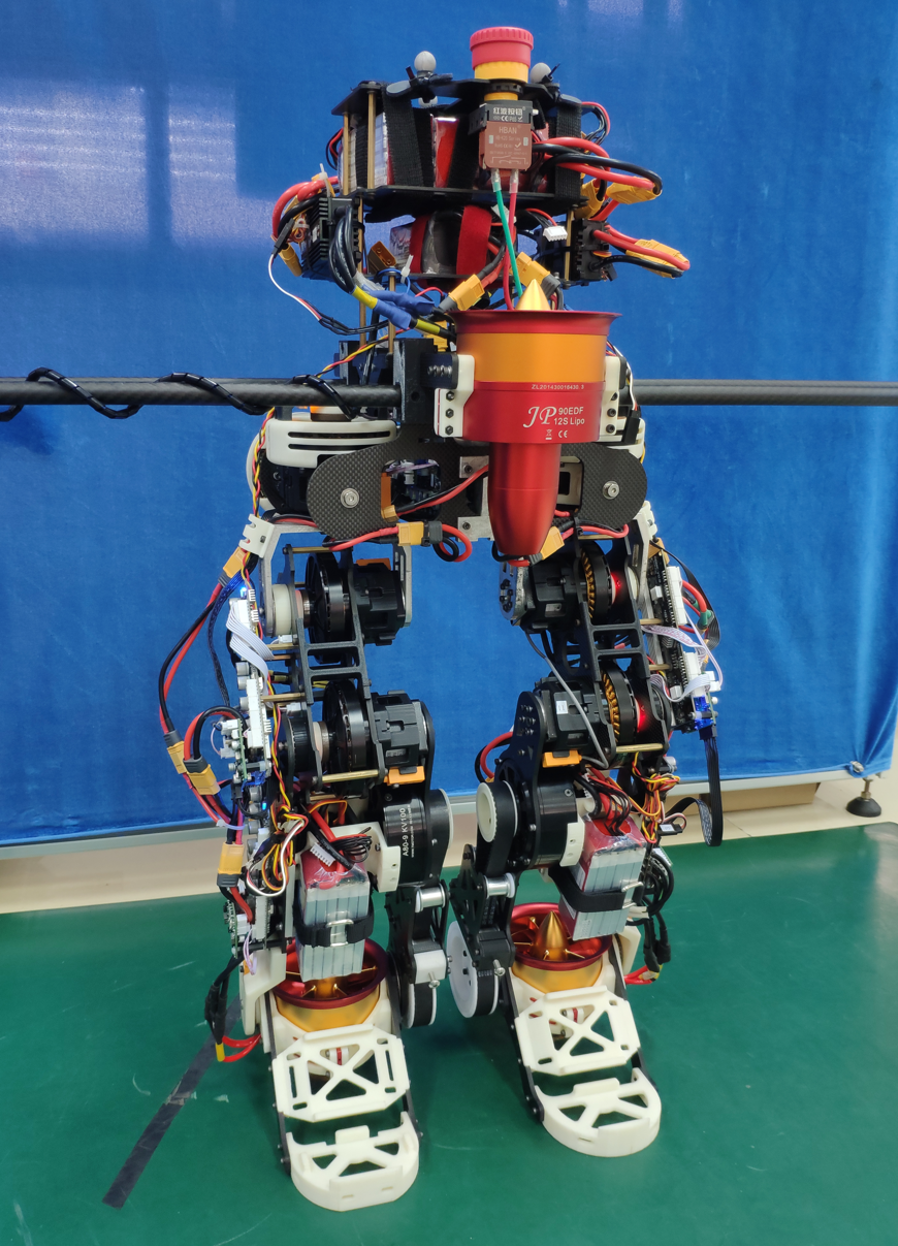}}
\caption{Jet-HR2: a flying humanoid robot based on ducted fans}
\label{fig:Fig1}
\vspace{-3.0mm}
\end{figure}

In relation to achieving good jumping performance, many methods were proposed owing to advances in actuator technologies, algorithms, and other optimizations. One approach is weight saving. Kojima \emph{et al.} proposed a high-stiffness optimized mechanical structure design, and its effectiveness was proven by a prototype robot JAXON3-P that dynamically jumped 300 mm in height \cite{kojima2019robot, kojima2020drive}. Compared with previous JAXON series, this new structural optimization successfully reduced the frame weight by approximately 62\%. Cassie \cite{xiong2018bipedal} was designed with a concentrated mass at the pelvis, and it has lightweight legs with leaf springs and a closed kinematic chain mechanism that can jump approximately 180 mm in height. Another approach is to enhance the power where pneumatic artificial muscle (PAM) actuators were introduced \cite{tondu2000modeling, nishikawa2014musculoskeletal}. Taking advantage of the properties of PAM in follow-up studies \cite{sulistyoutomo2018sequential, liu2018using, kaneko2016force}, the latest musculoskeletal humanoid could accomplish sequential jumping-stepping motions. Otani \emph{et al.} \cite{otani2018jumping} proposed a jump method by combining active joint driving with spring behavior in an actuator to achieve countermovement jumping motion. As a representative humanoid robot, ATLAS \cite{ATLAS2017} can perform standing long jump and execute complex tasks driven by hydraulic actuators. 

Although state-of-the-art miniature and multi-legged robots, such as JumpRoACH \cite{jung2016integrated}, Stanford Doggo \cite{kau2019stanford}, and Solo \cite{grimminger2020open} could jump multiple times higher than their size, real-sized humanoid robots in previous studies could not jump higher or farther than their own body length. Even the latest representative humanoid robot ATLAS \cite{ATLAS2017} could not jump higher than its height of 1.5 m. The performance of humanoid robots is still not up to the human level, especially with an increase in mass. On the other hand, even at the human level, robots may appear helpless on loose, collapse-prone, or cliff-like terrain.
\begin{figure}[t]
\setlength{\abovecaptionskip}{0.cm}
\setlength{\belowcaptionskip}{-0.cm}
\centerline{\includegraphics[width=0.5\textwidth]{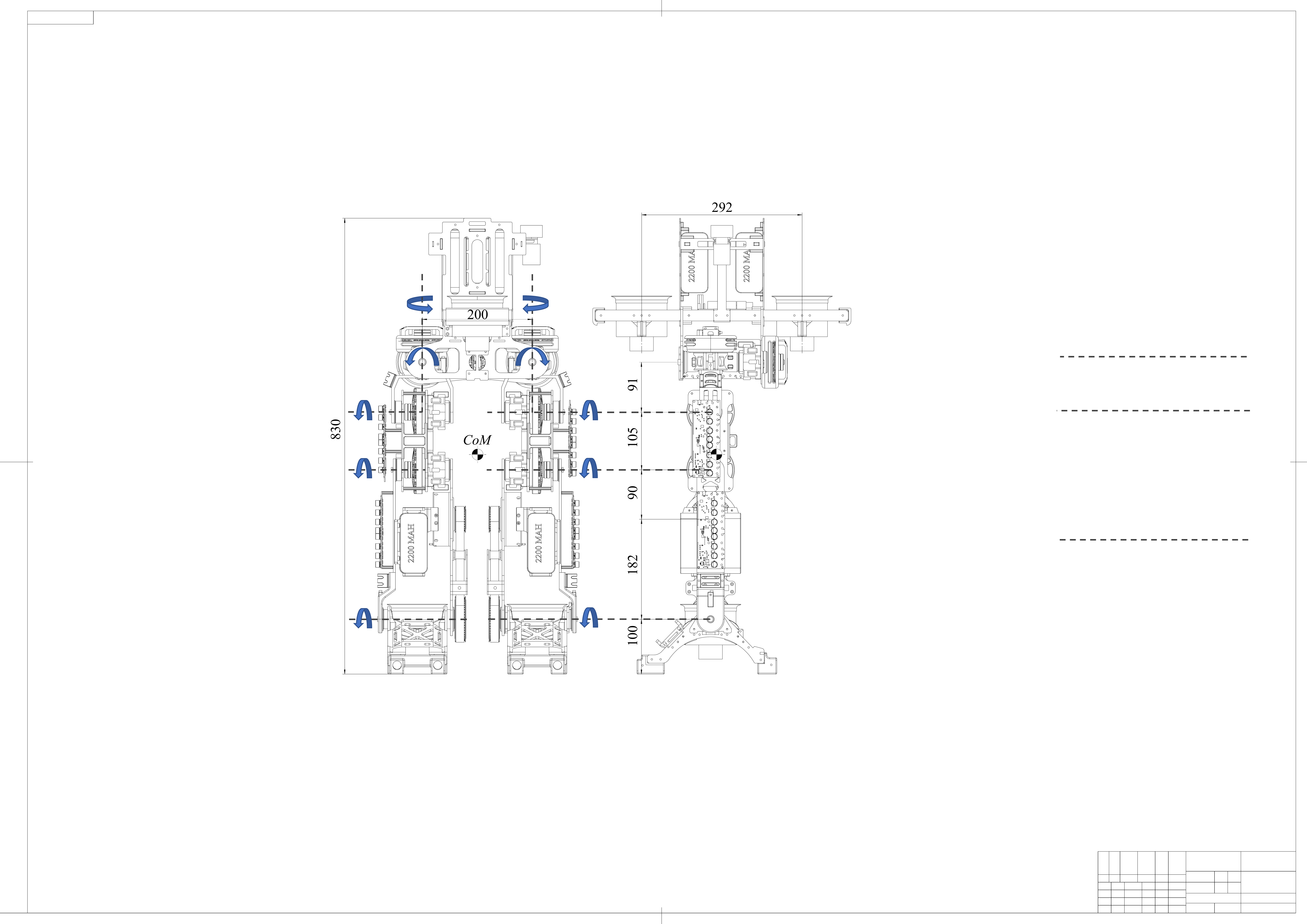}}
\caption{Dimensions and ducted fan configuration of Jet-HR2}
\label{fig:Fig2}
\end{figure}
This seems to be a limitation of using purely joint actuators to generate force. Besides, the long jumping capability has not been discussed for most of the above humanoid robots. Meanwhile, the landing impact should be considered for humanoid robots in many cases, which may increase the possibility of terrain collapse.

There are some studies for introducing thruster and aerodynamic lift to improve the dynamic performance of the traditional robot. Zhao \emph{et al.} \cite{zhao2018design} developed a dragon-like aerial robot based on thruster vectoring. The robot can transform in the air to fly across a narrow space by controlling gimballed pairs of ducted fans to adjust the thrust direction of each joint. Utilizing torsional spring, thruster, and flywheel, Salti-1P \cite{haldane2017repetitive} could reach high vertical jumping agility at 2.2 m/s with a height more than 8 times its size. Researchers from IIT had a complete flying simulation on iCub \cite{pucci2017momentum}, focusing on the control strategy and stabilization using jet turbines as thrusters. Our previous study \cite{huang2020three} proposed a jet-powered humanoid robot, Jet-HR1, with a ducted fan installed on each foot. By maintaining a quasi-static balance, the robot could step over a ditch with a span of 450 mm (as much as 97\% of the robot’s leg) in 3D stepping. These studies demonstrated the possibility of using thrust to enhance the locomotion of humanoid robots. However, to the best of our knowledge, no life-sized humanoid robot has yet achieved flight. In this study, a novel humanoid robot that can fly using a ducted fan propulsion system was developed to explore its potential value for search and rescue in complex environments.

The primary design and control challenge is to achieve the attitude stability of the robot in the air under low thrust-to-weight conditions. The design of the robot requires tradeoffs in terms of the mass distribution between the joint actuators and propulsion systems. The robot needs to ensure that the joints have sufficient power to perform tasks such as walking, but also the propulsion can meet the requirements of short flight distances and does not place excessive weight burden on the walking task. In addition, sufficient joint torque is a guarantee for adjusting the orientation when the thrust is delivered at full load. It is difficult to achieve a thrust-to-weight ratio of more than 2 or 4 for robots to meet these requirements, as is the case with conventional quadcopters. 

To overcome this challenge, the robot uses controlled thrust vectoring on the feet during flight to stabilize the attitude. The main advantage of this method is that it can provide more thrust to resist gravity compared with the method that uses thrust difference to generate torque. During the flight, the direction of the thrust is actively controlled by adjusting the attitude of the feet. With this method, the dive and spin motions caused by errors in the center-of-mass distribution and joint angle during the flight of the robot are effectively suppressed.

The remainder of this paper is organized as follows: Section \ref{sec:Mechanical Design} introduces the mechanical design and specifications. Section \ref{sec:Model and Control Strategies} presents the model and control strategy. The results of the takeoff experiments are presented in Section \ref{sec:Experiment}. Finally, Section \ref{sec:Conclusion} concludes the study.

\section{Mechanical Design}
\label{sec:Mechanical Design}
\subsection{Specifications of Jet-HR2}

Table. \ref{table:1} shows the main specifications of the prototype Jet-HR2. The robot has 10 degrees of freedom driven by brushless motor modules for ground locomotion and for adjusting the direction of the ducted fans installed in the feet during the flight. The propulsion system includes four ducted fans (Fig. \ref{fig:Fig2}). In addition to the two fans mounted on the palms of the feet, there are two other fans mounted at the front and rear of the waist to provide auxiliary thrust. To keep the robot lightweight, the main parts of the frame used the material of carbon fiber and ABS.
\begin{figure*}[t]
\vspace{2mm}
\setlength{\abovecaptionskip}{0.cm}
  \centering
    {\includegraphics[width=0.8\textwidth]{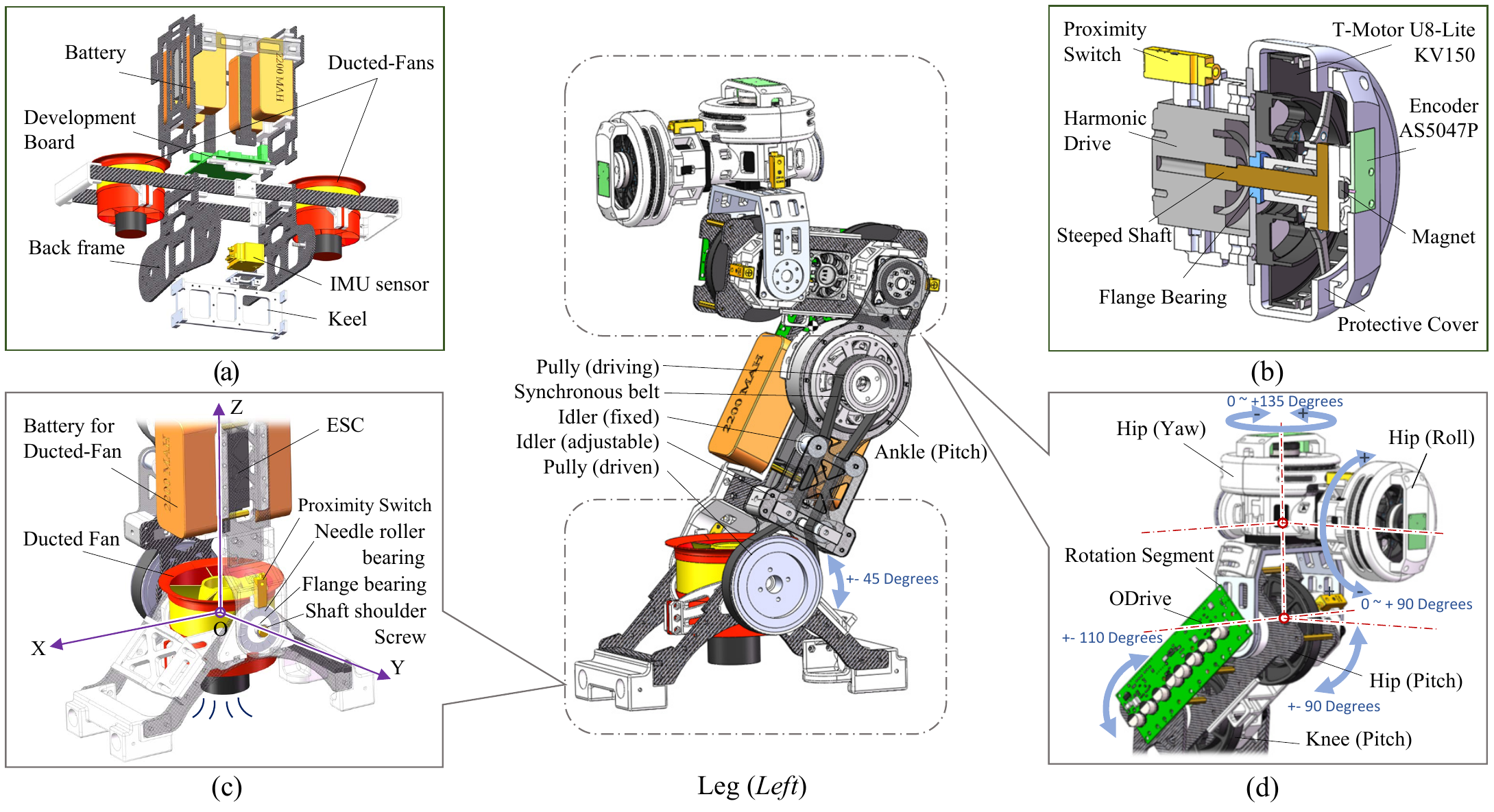}} \\
  \caption{Assembly of mechanical components of the robot leg: (a) Waist (b) Modular joint utilized in hip and knee (c) Feet (d) Thigh}
	\label{fig:Fig3}
\end{figure*}
\bgroup
\def\arraystretch{1.3}
\begin{table}[b]\footnotesize    
\setlength{\abovecaptionskip}{-0.0cm}
\centering
\caption{Main Specifications of JET-HR2}
\label{table:1}
\begin{tabular}{p{3.15cm} p{4.6cm}}  
\Xhline{2\arrayrulewidth}
\hline  
\hline  
Component                               & Description                                                                                       \\ \hline 
Mass (with battery)                     & 17kg                                                                                              \\
Height                                  & 830mm                                                                                             \\
Length of leg                           & 480mm                                                                                             \\
Maximum total thrust                    & 200N (Ducted fan x4)                                                                              \\
Ducted fan                              & {$\Phi$} 90mm, 48V, 488g \newline Max thrust 50N@90A                                                 \\
Actuators (except ankle)                & T-Motor Co. Ltd. U8-Lite KV150 24 V, 1.83 Nm; A-80-9, 24V, 18 Nm, 2573rpm                         \\
Reducer (except ankle)                  & Harmonic Drive \newline CSF-11-50-1U-CC-SP, 50:1                                                 \\
Actuator of ankle joints                & T-Motor Co. Ltd. A80-9 \newline Reduction ratio 1:9, 24V \newline Normal/Peak torque: 9Nm/ 18Nm \newline Normal speed: 245rpm \\
Driver of actuator                      & ODrive, ODrive Robotics Co. Ltd. \newline Ver 3.5, Can Bus                                        \\
Ankle joint reduction ratio             & 28:50                                                                                             \\
Battery of motor                        & 6S, 22.2V, 1300 mAh                                                                               \\
Battery of ducted-fan                   & 12S, 44.4V, 2200 mAh                                                                               \\
IMU                                     & ASENSING CO. Ltd. INS550 \newline 250 Hz, Resolution 0.3                                          \\
\hline  
\hline  
\end{tabular}

\end{table}
\subsection{Modular Joint Design}

Jet-HR2 uses modular joints (Fig. \ref{fig:Fig3}b) for hip and knee degrees of freedom. The utilization of modular joints on robots has several advantages, including easier repair, modification, and a simplified design \cite{bledt2018cheetah, katz2019mini}. The coordination of high-torque-density brushless motors and a harmonic drive can achieve a high output torque, accurate force/torque control precision and remains back-drivable, similar to most state-of-the-art legged robots. The modular joint in Jet-HR2 is composed of a brushless motor (T-Motor U8-Lite KV150) and a harmonic drive (CSF-11-50-1U-CC-SP). Here, we chose a high transmission ratio of 50, in which the maximum effective torque reached 30 Nm. The torque enables the robot to adjust the direction of the ducted fan installed in the feet, even when the legs are fully extended horizontally with the maximum thrust output. The estimated maximum torque caused by the ducted fan was 24 Nm. Furthermore, we integrated an absolute magnetic rotary encoder (AS5047P) placed off-axis on the protective cover to communicate with the motor controller ODrive (ODrive Robotics, Richmond, CA) based on CAN-Bus. For the initial position correction, the proximity switch is settled at the harmonic drive set printed in ABS plastic materials. In addition to high torque density and high control precision, the modular joint is compact and weighs 604 g.
\subsection{Leg Design}
To reduce the leg inertia, which is related to the placement of the knee actuators, many optimizations have been made by previous studies, such as the Cassie \cite{xiong2018bipedal}, Minitaur \cite{kenneally2016design}, and MIT Cheetah series \cite{nguyen2019optimized, park2015variable}. These robots have the knee joint placed close to the hip joint to place the center of mass (CoM) on the higher part of the robot as much as possible. Similarly, we optimized the leg design on Jet-HR2 by aggregating the hip joint (pitch) and knee joint on the same carbon frame as integrated. Fig. \ref{fig:Fig3}d shows the mechanical design of the thigh part of the robot. The three-DoF hip joints are allocated separately, with the hip joint (roll) mounted on the waist. The hip joint (yaw) is located between the two carbon frames of the pelvis. In addition to reducing leg inertia, the roll-yaw-pitch configuration enables the robot to have a maximum range of motion. The hip joint (roll) could rotate 0–90°, the hip joint (yaw) could rotate 0–135°, the hip joint (pitch) ± 90°, and the knee joint ± 110°. This range of workspace could allow the robot to use its whole leg length in dynamic motions, such as stepping over a large ditch in 2D gaits \cite{liu2018jet}, which contributes to enhancing the locomotion performance. The actuators on the shank are different from those of the thigh parts mentioned above. Here, we used the integrated actuator A80-9 motor (T-Motor Co. Ltd.) to satisfy the higher speed requirement for thrust vector control. We evaluated the velocity and torque on a physical simulation platform, PyBullet \cite{coumans2021}, to determine the appropriate transmission ratio for the ankle joints. The torque is transmitted to the feet of the robot through a 28:50 belt drive. The belt transmission further contributes to reducing leg inertia and helps improve the stiffness.
\subsection{Waist Design}
The primary function of the waist frame (Fig. \ref{fig:Fig3}a) is to support two ducted fans and the module of roll hip joints. To save weight, the frame is mainly made of carbon fiber. The basic frame consists of an aluminum keel joining the front and back boards. The modules of the roll hip joint are mounted on the backboard to drive the entire thigh to lateral rotation. In addition, bearings are embedded in the front board, and the front of the thighs is attached by shoulder screws for support. Two ducted fans are fixed in the waist’s front and rear, respectively. To protect the ducted fans when the robot falls down, crash bars made by ABS are installed in front of the ducted fan. The IMU sensor and battery are also fixed on the waist frame.
\section{Model and Control Strategies}
\label{sec:Model and Control Strategies}
The following challenges affect the stability of the takeoff attitude. First, the robot is not perfectly centrosymmetric. The center of mass varies in the sagittal plane as the robot squats, leans forward, or leans back (Fig. \ref{fig:Fig4}). The difference in the force arms of the individual thrusters at the center of mass causes the robot to dive or lean back in the pitch direction. Second, owing to the interference of the ground, the thruster of the robot’s feet is not free to adjust its orientation until the feet are completely off the ground. As a result, the takeoff posture and thrust distribution should be carefully considered. Third, owing to the end thrusts, there are position control errors in the joints, which cause the robot to spin in the yaw direction because of force couples generated by errors in the direction of the thrusts.
\subsection{Analysis of Force Model}
The model is established to analyze the forces during takeoff in a fixed position and how the robot’s attitude can be effectively adjusted in the air under low thrust-to-weight ratio conditions. The related parameters of the simplified model are listed in Table \ref{table:2}. The following assumptions are made to simplify the robot model for targeted solutions to the problems of dive and yaw rotation during takeoff: 1) the robot maintains a fixed position during takeoff and is considered as a rigid body and 2) the robot legs remain parallel to each other. In other words, the components of the robot’s feet in the $z$-axis and $x$-axis remain the same in the body frame. 3) The pitch attitude of the ducted fans can be independently controlled in the body frame.
\bgroup
\def\arraystretch{1.3}
\begin{table}[!b]\footnotesize 
\centering
\caption{Main Variables of The Model and Methods}
\label{table:2}
\begin{tabular}{p{2.3cm} p{5.2cm}}
\Xhline{2\arrayrulewidth}
\hline  
\hline  
Symbol                                                      & Description                                                                                        \\ \hline 
$M$                                                         & Total mass of robot                                                                                \\
${ }^{w} \boldsymbol{F}_{\text {all }}$                     & Scalar {$\boldsymbol{a}$} and its first order difference                                           \\
${ }^{w} \boldsymbol{\tau}_{\text {all }}$                  & Vector {$\overrightarrow{\boldsymbol{a}}$} and its components                                      \\
$\theta_{R}, \theta_{L}$                                    & Pitch angles of right and left foot with the reference of the x-axis in body frame $\{\boldsymbol{B}\}$ \\
$\theta_{\text {pitch }}$                                   & Pitch angle of the robot’s waist link in world frame $\{\boldsymbol{W}\}$                          \\
$\operatorname{Rot}\left(y, \theta_{\text {pitch }}\right)$ & Rotation matrix constructed by pitch world frame $\{\boldsymbol{W}\}$                              \\
$f_{F}, f_{B}, f_{R}, f_{L}$                                & Thrust of ducted-fans installed in waist and feet respectively.                                    \\
$\left[x_{c}, y_{c}, z_{c}\right]$                          & Position of center of mass in $\{\boldsymbol{B}\}$                                                 \\
$\left[p_{f x},-1 / 2 L_{f}, p_{f z}\right]$                & Position of left foot’s ducted-fan in $\{\boldsymbol{B}\}$                                         \\
$\left[p_{f x}, 1 / 2 L_{f}, p_{f z}\right]$                & Position of right foot’s ducted-fan in $\{\boldsymbol{B}\}$                                        \\
$L$                                                         & Distance between front and back ducted-fan                                                         \\
$L_{f}$                                                     & Distance between feet’s ducted fan                                                                 \\
\hline  
\hline  
\end{tabular}
\end{table}  
\begin{figure}[!t]
\vspace{2.5mm}
\setlength{\abovecaptionskip}{0.0cm}
\centerline{\includegraphics[width=0.47\textwidth]{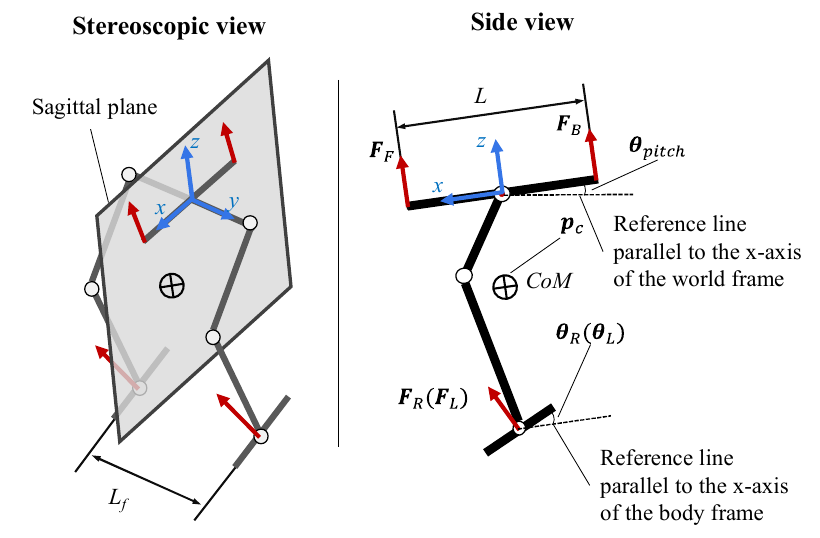}}
\caption{Simplified model}
\label{fig:Fig4}
\vspace{-3.0mm}
\end{figure}

Based on these assumptions, the robot becomes a 2D model in the sagittal plane. However, the feet are symmetrically distributed on both sides of the sagittal plane. Using forward kinematics, the model is expressed by the following two equations:

{\small
\begin{equation}
\label{eq:1}
{}^{w} \!\boldsymbol{F}_{a l l}\!=\! \operatorname{Rot}\!\left(y, \theta_{\text {pitch }}\!\right)\!\!\left[\!\!\!\!\begin{array}{c}f_{L} \sin \theta_{L}\!+\!f_{R} \sin \theta_{R} \\ 0 \\ f_{B}\!+\!f_{F}\!+\!f_{L} \cos \theta_{L}\!+\!f_{R} \cos \theta_{R}\end{array}\!\!\!\!\right]\!\!-\!\!\left[\!\!\!\!\begin{array}{c}0 \\ 0 \\ M g\end{array}\!\!\!\!\right]
\end{equation}

\begin{equation}
\label{eq:2}
{ }^{w} \boldsymbol{\tau}_{\text {all }}\!=\!\operatorname{Rot}\!\left(y, \theta_{\text {pitch }}\!\right)\!\!\left[\!\!\!\!\begin{array}{c}\frac{1}{2} L_{f}\left(f_{L} \cos \theta_{L}\!-\!f_{R} \cos \theta_{R}\right) \\ T_{y_{-} 1}\!+\!T_{y_{-} 2}\!+\!T_{y_{-} 3} \\ \frac{1}{2} L_{f}\!\left(f_{R} \sin \theta_{R}\!-\!f_{L} \sin \theta_{L}\right)\end{array}\!\!\!\!\right]
\end{equation}
where
\begin{equation}
\label{eq:3}
\left\{\!\!\!\begin{array}{l}T_{y_{-} 1}\!=\!f_{B}\left(L / 2\!+\!x_{c}\right)\!-\!f_{F}\left(L / 2\!-\!x_{c}\right) \\ T_{y_{-} 2}\!=\!\left(f_{L} \cos \theta_{L}\!+\!f_{R} \cos \theta_{R}\right)\left(x_{c}-p_{f x}\right) \\ T_{y_{-} 3}\!=\!-\!\left(f_{L} \sin \theta_{L}\!+\!f_{R} \sin \theta_{R}\right)\left(z_{c}\!-\!p_{f}\right)\end{array}\right.
\end{equation}
 }

Equation \eqref{eq:1} shows that all the thrusters contribute to the vertical thrust. As the pitch angle $\theta_{\text {pitch }}$  increases, some vertical thrust is lost and converted to horizontal thrust, especially because of the two fans at the waist. As a result, under conditions of a low thrust margin, the waist link should be kept as horizontal as possible during takeoff to provide vertical thrust. In addition, proper use of waist inclination can effectively counteract the horizontal thrust due to foot inclination and avoid drifting in the front or back direction during takeoff.

Equation \eqref{eq:2} and Equation \eqref{eq:3} describe the influence of the position of the robot’s CoM on the takeoff process. The influence is mainly in the pitch direction. Here, $T_{y_{-} 1}$ indicates the torque in the $y$-direction generated by the ducted fans on the waist. To avoid generating external torque in the $y$-direction and make the robot dive down, the thrust difference between the front and back fan output is required once $x_{c}$  is not 0. Thus, the fan at the waist does not achieve the full maximum output, resulting in a partial loss of thrust. $T_{y_{-} 2}$ shows the torque due to the difference in the position of the ducted fan of the feet from the CoM in the $x$-direction. This item cannot be eliminated completely by controlling the thrust output. $T_{y_{-} 3}$ is caused by the position difference in the $z$-axis between the foot’s ducted fans and the CoM. Owing to the mass distribution, this difference is significant compared with that on the $x$-axis. Even a small horizontal thrust component in the feet can produce a significant torque. The component of $T_{y_{-} 3}$ shows the advantages of using thrust vector control on feet’s ducted fan to control the robot’s attitude in the pitch direction.

Fig. \ref{fig:Fig5} compares the results of the two strategies to the general torque in the pitch direction. One is using the thrust differential (DT) between the front and rear ducted fans on the waist, similar to a multi-axis vehicle.
The other is using thrust vector control (TVC) on the feet to generate pitch torque. The maximum torque that the robot can obtain at different pitch angles was used as an evaluation indicator to assess the two methods. Because of the constraint that the vertical thrust should be at least equal to the gravity, both the thrust difference or the horizontal thrust of the feet cannot achieve the maximum thrust output. Three different takeoff postures (detailed in Fig. \ref{fig:Fig6} and Table \ref{table:3}) were compared. In all postures, the method of TVC’s boundaries of pitch torque is three times that of DT in both clockwise and counterclockwise directions. This result is mainly due to the considerable length of the force arm $\left(z_{c}-p_{f z}\right)$ between CoM and the foot. The DT method requires three times the current distance between the waist fans if the same value is to be achieved. However, this would make the robot footprint too large and unrealistic.

Equation \eqref{eq:2} also describes that the ducted fans on the feet can generate torque in the yaw and roll directions by adjusting the output and attitude difference, respectively. The control of these torques is simple because it is independent of the position of the CoM.
\bgroup
\def\arraystretch{1.3}%
\begin{table}[!b]\footnotesize  
\centering
\caption{Parameters of Different Take-off Posture}
\label{table:3}
\begin{tabular}{m{0.7cm} m{2.1cm} m{2.1cm} m{2cm}}  
\Xhline{2\arrayrulewidth}
\hline
\hline
Posture  &  Position of CoM \newline $\left(x_{c}, z_{c}\right)$ (mm) & Position of feet $\left(p_{f x}, p_{f z}\right)$ (mm) & Range of feet’s pitch angle $\left(^{\circ}\right)$   \\ \hline 
$P_{1}$   & (25, -243) & (20, -610) & (-74, 90) \\
$P_{2}$   & (20, -265) & (10, -650) & (-90, 90) \\  
$P_{3}$   & (50, -225) & (70, -580) & (-82, 90) \\
\hline  
\hline  
\end{tabular}
\end{table}
\begin{figure}[!t]
\vspace{1.0mm}
\setlength{\abovecaptionskip}{0.cm} 
\setlength{\belowcaptionskip}{-0.cm}
\centerline{\includegraphics[width=0.5\textwidth]{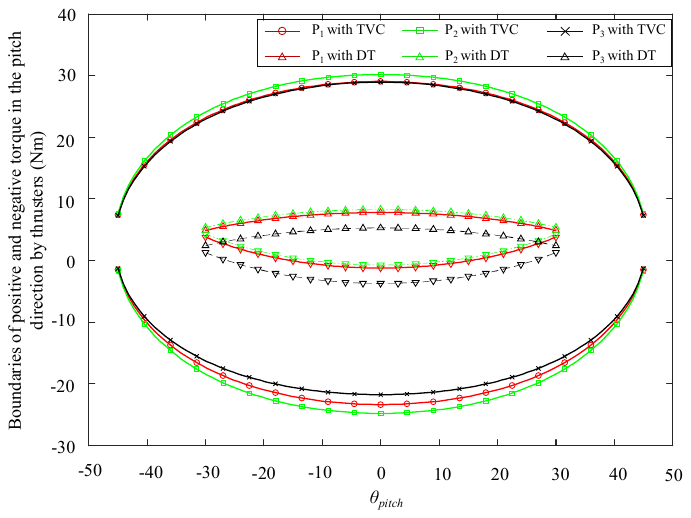}}
\caption{ Comparison of torque boundaries in pitch direction under the condition that the vertical thrust is greater than or equal to the gravitatio}
\label{fig:Fig5}
\vspace{-2.0mm}
\end{figure}
\begin{figure}[!t]
\setlength{\abovecaptionskip}{0.cm}
\setlength{\belowcaptionskip}{-0.cm}
\centerline{\includegraphics[width=0.5\textwidth]{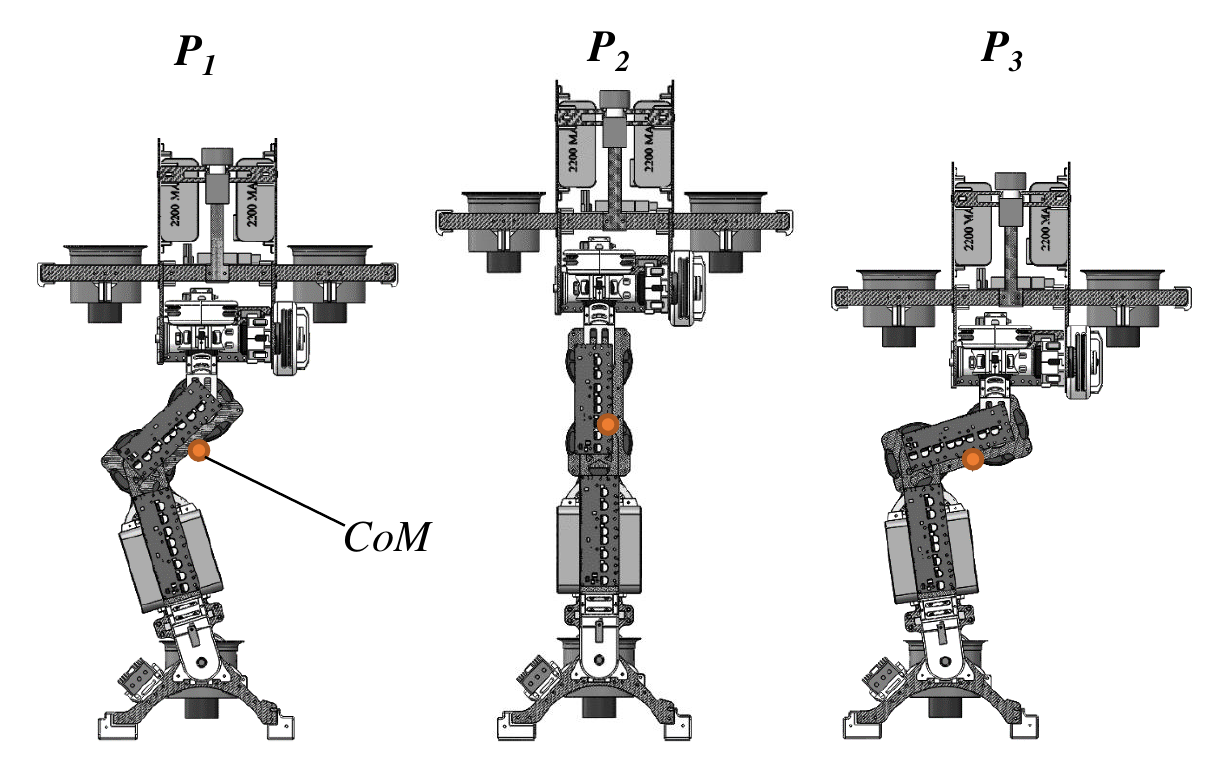}}
\caption{Comparison of three take-off postures}
\label{fig:Fig6}
\vspace{-2.0mm} 
\end{figure}
\begin{figure}[!t]
\setlength{\abovecaptionskip}{0.cm}
\setlength{\belowcaptionskip}{-0.cm}
\centerline{\includegraphics[width=0.5\textwidth]{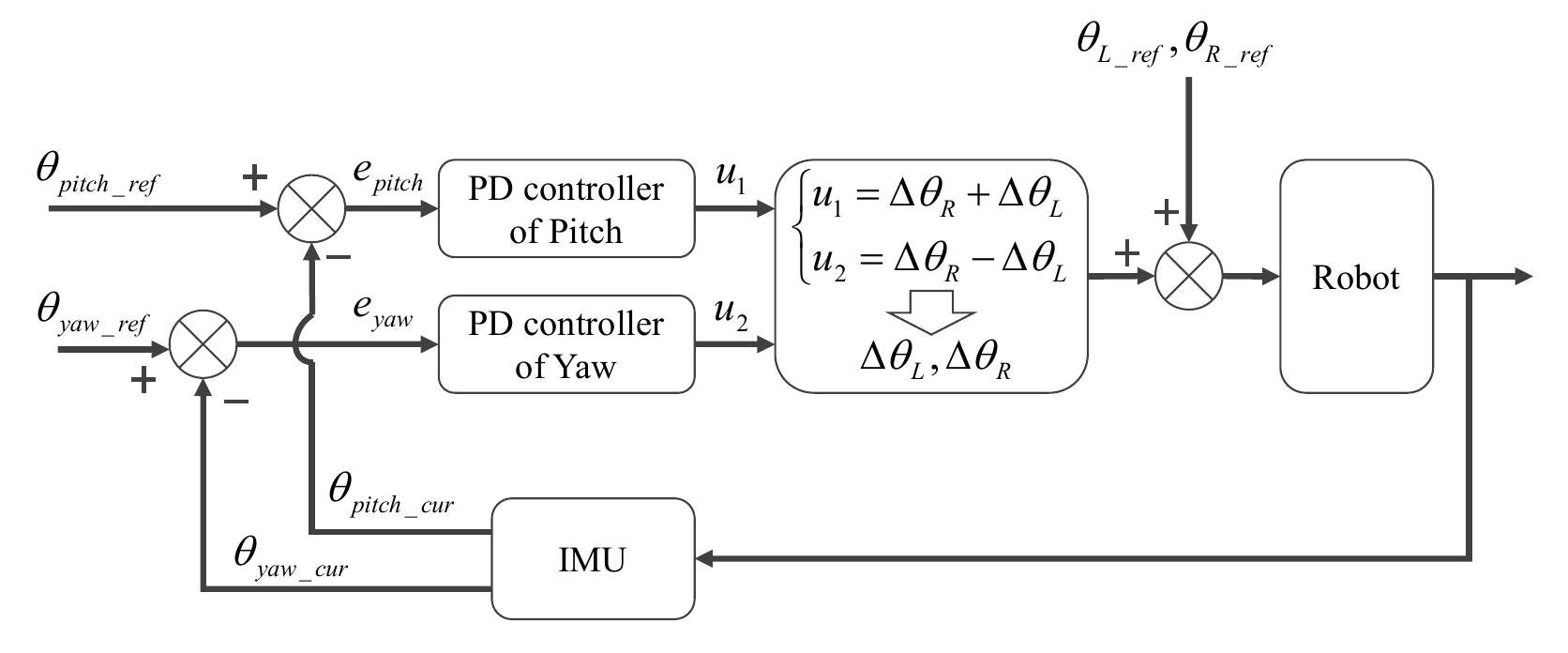}}
\caption{Frame of control strategies}
\label{fig:Fig7}
\end{figure}
\subsection{Control Strategies}
A control strategy based on thrust vector control is proposed to enable the robot to take off with a stable attitude. The strategy is primarily designed to suppress dives and spins. According to the analysis above, the strategy controls the pitch angle of the ducted fan of the feet rather than the magnitude of the thrust. The output of each thrust is set to be equal and preplanned according to the required acceleration of the takeoff. The strategy considers two aspects. First, the horizontal thrust generated by the control of the pitch angle of the feet is used to create a moment to control the pitch angle of the robot and avoid diving. Second, the pitch angle difference between the two feet is controlled to create a yaw moment to suppress rotation. All controls are based on the PD controller (Fig. \ref{fig:Fig7}).
\section{Experiment}
\label{sec:Experiment}
Takeoff experiments were conducted to examine the performance of the proposed thrust vector control on the robot’s feet. The purpose was as follows: 1) to check whether the propulsion system can enable the robot to take off and 2) to examine whether control strategies can effectively suppress dive and spin movements during takeoff. To clearly show the effect, three conditions are compared: 1) pitch and yaw controllers are put on, 2) only the pitch controller is put on, and 3) all controllers are put off. All trials used the same posture to take off. To prevent the robot from losing control and falling down, two carbon tubes were mounted on the robot’s waist to catch it. 

The robot performed three takeoffs under each condition. Fig. \ref{fig:Fig8} compare the results of the robot’s motion sequences. In all trials, the propulsion system successfully enabled the robot to take off completely. According to the results, the effect of the proposed control strategy is significant. The movement of dive and spin is effectively suppressed when both pitch and yaw controllers are applied. With the controller, the robot reached a height of 1000 mm in 2 s (Fig. \ref{fig:Fig8}a). Without a pitch controller, the robot rapidly dived forward approximately 1 s after takeoff (Fig. \ref{fig:Fig8}b), and without yaw control, the robot spun during takeoff, reaching more than 90° within 1 s (Fig. \ref{fig:Fig8}c).

\begin{figure}[!t]
\vspace{0.2cm}
\setlength{\abovecaptionskip}{-0.15cm}
\centerline{\includegraphics[width=0.46\textwidth]{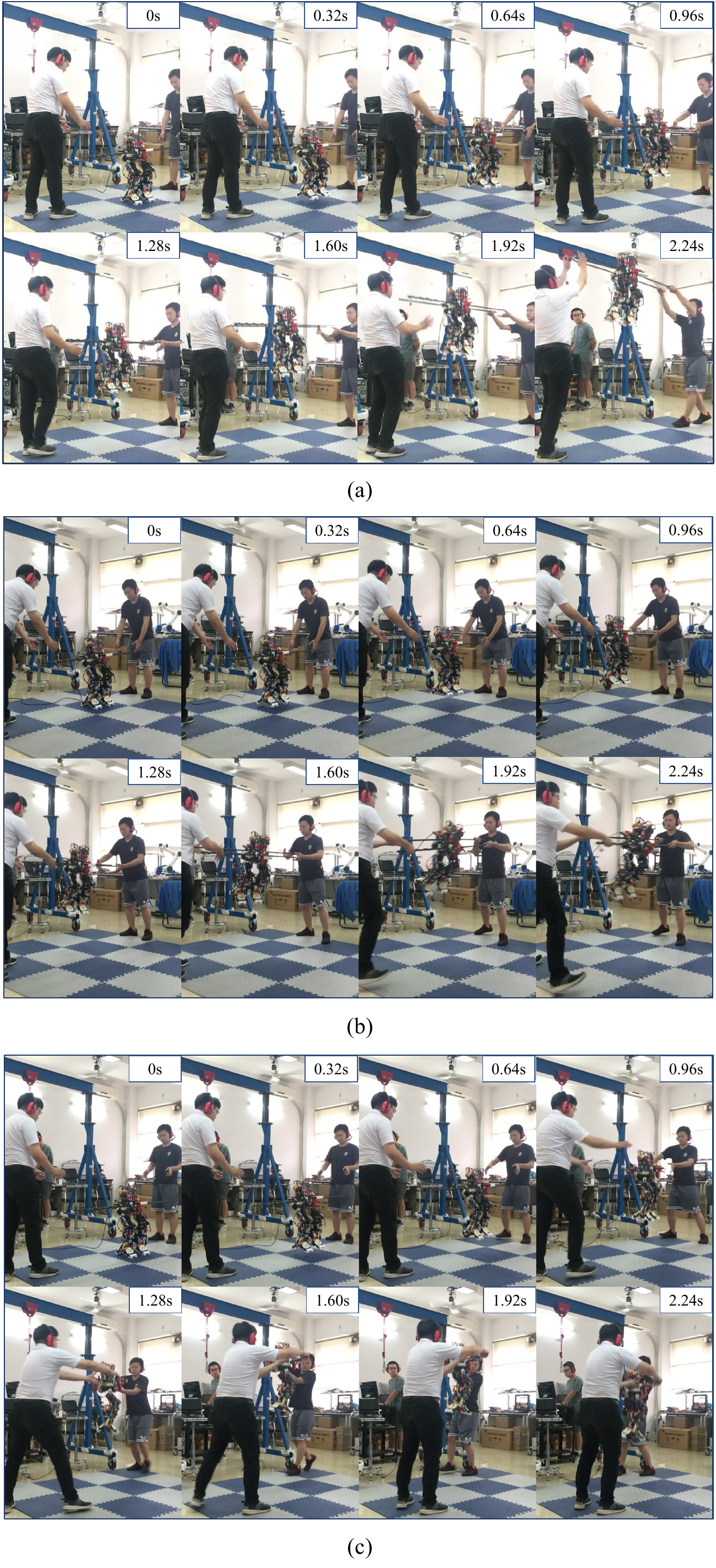}}
\caption{Experiment result of different control strategies, in which (a) Taking off with pitch and yaw controllers; (b) Taking off without controller; (c) Taking off with only pitch controller}
\label{fig:Fig8}
\end{figure}

Fig. \ref{fig:Fig9} shows the comparison of the robot’s attitudes under three conditions. Fig. \ref{fig:Fig9}a and Fig. \ref{fig:Fig9}b reveal that the controllers behave stably in the pitch direction with high repeatability of the pitch attitude in three takeoffs. When both controllers are on, the robot briefly tilted backward about 1° after takeoff and quickly adjusted to the opposite direction. During 1.5-2.2 s,  the pitch angle stabilized at around -5°. As a comparison, when both controllers are off, the robot briefly tilted backward after takeoff, then rapidly dived forward and finally lost control with a large pitch angle at around 30° (Fig. \ref{fig:Fig9}c). The performance of the controllers for the yaw angle is not ideal. Although the spin behavior was effectively suppressed, oscillations (around ± 13°) were observed (Fig. \ref{fig:Fig9}d). As a comparison, when the yaw controller is off, the robot rapidly turned to the negative yaw direction and exceeded 40° (Fig. \ref{fig:Fig9}e).
\begin{figure*}[t]
\setlength{\abovecaptionskip}{0.cm}
  \centering
    {\includegraphics[width=1\textwidth]{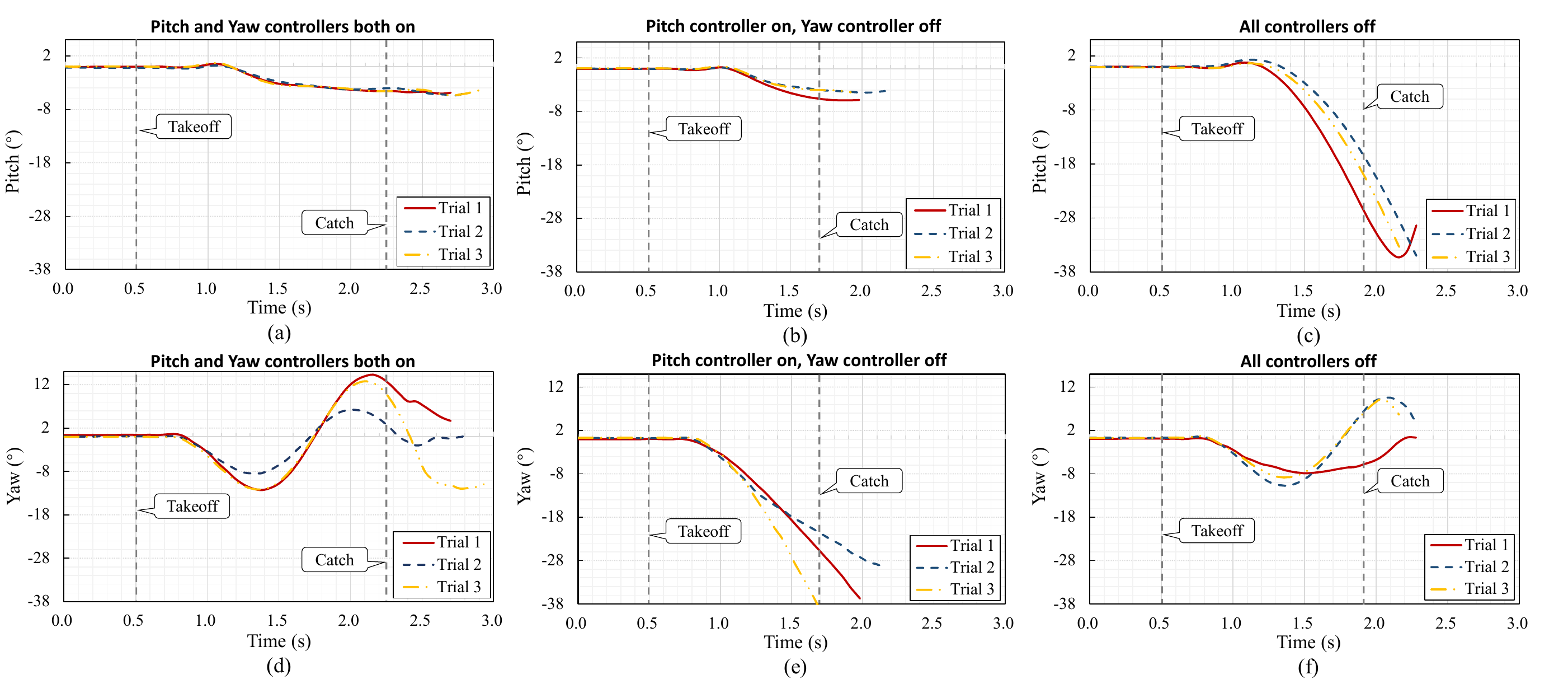}} \\
  \caption{Comparison of robot’s attitude during taking off under three conditions}
	\label{fig:Fig9}
\end{figure*}

Fig. \ref{fig:Fig10} describes the controller’s output on the robot’s feet corresponding to the condition when both controllers are put on (Fig. \ref{fig:Fig9}a and Fig. \ref{fig:Fig9}d). From the curve of the feet angle (Fig. \ref{fig:Fig10}a), it can be found that the left and right feet rotated in a different direction (about -9° and 1°) when the yaw angle decreased (1-1.4 s). This behavior created momentum to resist the movement in the yaw direction. When the yaw angle increased, feet rotated in the converse direction to counteract the effect. At 1.4 s and 2.1 s, both the yaw angle and the angle difference between the left and right feet are close to their peak. The summation average of the pitch angle of the left and right feet is calculated respectively to show the corresponding control output (Fig. \ref{fig:Fig10}b). According to Fig. \ref{fig:Fig10}b, the output of the controller and the pose performance of the robot are consistent (Fig. \ref{fig:Fig9}a). While the robot tilted back less than 1° at 1 s, the feet rotated about 2° to suppress it. After 1.3 s, the summation average angle of feet gradually stabilized near 25° with the pitch angle of the robot stabilized near 5°.

Several phenomena and issues remain to be noted and considered in the experimental results. There are steady-state errors in the pitch direction. This might be due to the configuration of the robot and the anterior-posterior asymmetry of the center of mass in the sagittal plane. Adding the integral term might improve the problem. It is worth noting that even with the presence of steady-state errors, the robot did not drift significantly back and forth during takeoff. This may indicate that the desired angle to maintain hover or takeoff is not perfectly horizontal due to the structure of the robot. Another phenomenon worth noting is that the robot dives forward rapidly when the controllers are all off, but there is no significant rotation in the yaw direction. An explanation of this phenomenon awaits a more specific kinetic analysis to follow. The experiments also showed that the roll direction does not tilt significantly because of the left-right symmetry of the robot, but there is a certain amount of drifting, which needs to be solved by adding further sensors and improving the controller.
\begin{figure}[t]
\vspace{2mm}
\setlength{\abovecaptionskip}{0.cm}
\setlength{\belowcaptionskip}{-0.cm}
\centerline{\includegraphics[width=0.5\textwidth]{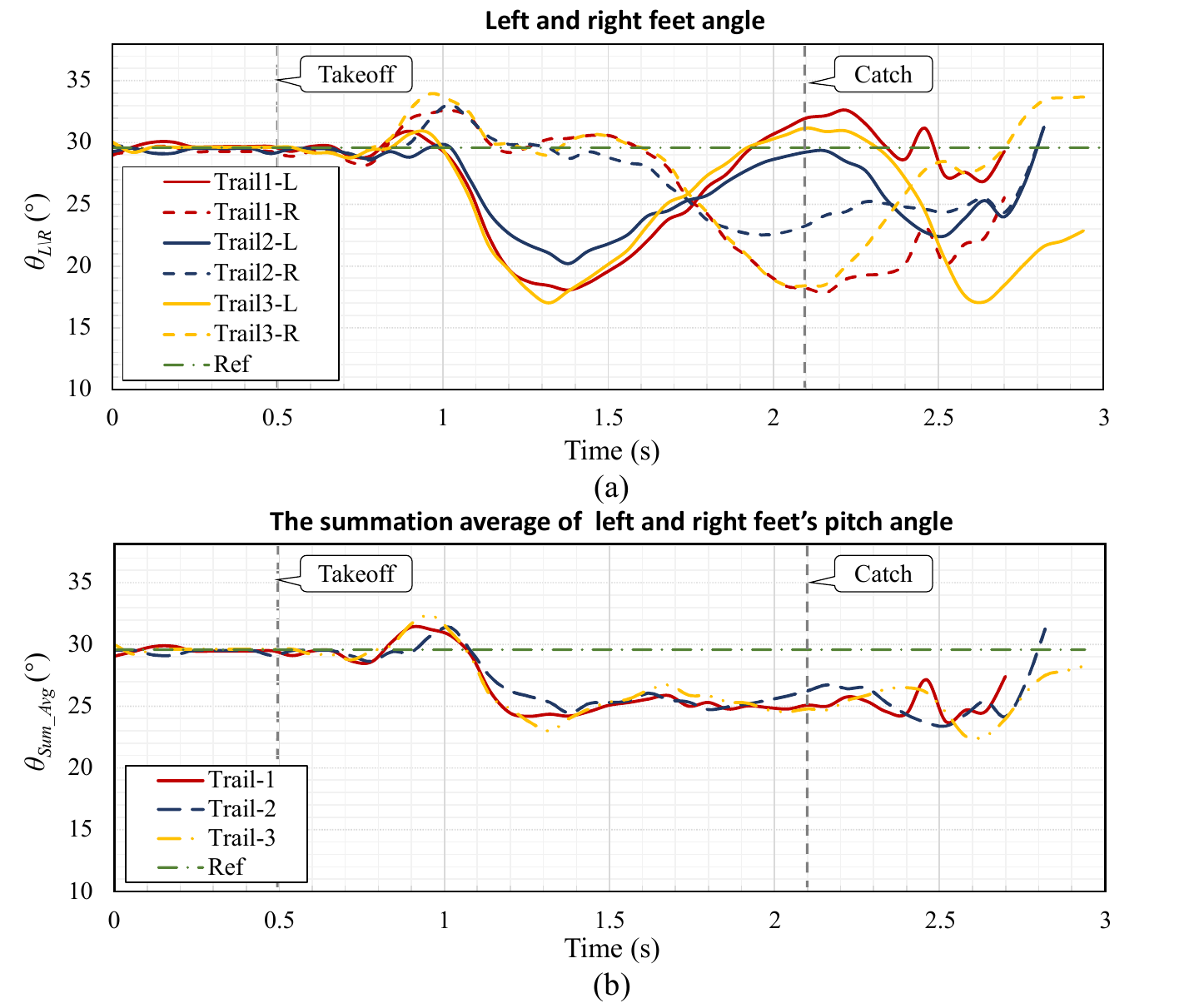}}
\caption{Comparison of robot’s foot angle during taking off when both controllers are on}
\label{fig:Fig10}
\end{figure}

\section{Conclusion}
\label{sec:Conclusion}
In this study, a flying humanoid robot based on the concept of thrust vector control was designed. A simplified force model was used to analyze how to maintain attitude stability during the robot’s takeoff. In the analysis, we also compared the two methods that generated the deflection torque in the pitch direction. The results indicate that using thrust vector control on the ducted fan of the feet can generate a considerable torque in the pitch direction. The results of the takeoff experiments with the prototype robot showed that the robot performed as expected. The robot achieved a successful takeoff under low thrust-to-weight ratio conditions. Moreover, the control strategy was effective in keeping the robot attitude stable during the takeoff. 

The main contributions of this study are summarized as follows:

1) Our newly designed humanoid robot system successfully achieved a fully controllable takeoff, which provides a new reference for research on flight methods for humanoid robots.

2) The control strategy of the TVC on the robot’s feet was proposed, and its effectiveness was experimentally proved. This provides a reference for the attitude stabilization methods under low thrust-to-weight ratio conditions.

\addtolength{\textheight}{-10.2cm}   
\bibliographystyle{IEEEtran}
\bibliography{bibliography}

\end{document}